# A Fast Feature Point Matching Algorithm Based on IMU Sensor

Lu Cao
School of Software and Microelectronics
Peking University
Beijing, China
andrewcao95@pku.edu.cn

*Abstract*—In simultaneous localization and mapping (SLAM), image feature point matching process consume a lot of time. The capacity of low-power systems such as embedded systems is almost limited. It is difficult to ensure the timely processing of each image information. To reduce time consuming when matching feature points in SLAM, an algorithm of using inertial measurement unit (IMU) to optimize the efficiency of image feature point matching is proposed. When matching two image feature points, the presented algorithm doesn't need to traverse the whole image for matching feature points, just around the predicted point within a small range traversal search to find matching feature points. After compared with the traditional algorithm, the experimental results show that this method has greatly reduced the consumption of image feature points matching time. All the conclusions will help research how to use the IMU optimize the efficiency of image feature point matching and improve the real-time performance in SLAM.

## I. INTRODUCTION

The localization technology based on RGB-D sensor become a hot research problem in recent years. RGB-D sensors (such as Kinect, Xtion) can provide color (RGB) and depth image information at the same time. Compared with ordinary camera, it has good continuity and abundant environmental information, so use RGB-D sensor in SLAM can get a higher positioning accuracy, and easier to reconstruct 3D map [1].

Localization algorithm based on RGB-D sensor has a lot of kinds. Such as Henry [2] and Endres [3] proposed RGB-D SLAM algorithm based on image matching technology, Kerl [4] and Audras [5] proposed intensive feature tracking method based on photometric model, Stuckler [6] proposed method based on point cloud matching technology. The RGB-D SLAM algorithm proposed by Endres using SIFT [7], SURF [8], ORB [9] algorithm for feature extraction and matching, and then use RANSAC algorithm [10] to get the initial pose estimation and remove outliers, using ICP point cloud matching algorithm [11] to improve the pose estimation. But the efficiency of the system is not high, frame dropping often appeared when a camera moving quickly. The system consumes a lot of time in the image feature point matching process, for example, when using SIFT feature matching algorithm, only the adjacent feature points matching two images need more than 42,000 seconds.

According to reduce time consume when matching feature points, Patrick presents an algorithm to reduce computational time by offloading the processing to a cloud comprised of a cluster of compute nodes [12]. Sturm present a 3D model [13] based tracker runs parallel with the MonoSLAM [14]. Wang took advantage of a modified SURF feature to estimate the state of the robot [15]. This paper proposes the following improvements: introducing the IMU sensor in the system to measure the value of angular velocity and acceleration of the camera between previous frame and current frame. Obtaining camera's attitude change and position deviation via integral calculation. Then calculate the rotation matrix R and displacement matrix t, and predict the location of the previous frame image feature points in the current frame image according to planar homography [16]. During feature point matching, don't need to traverse search the whole image, just traversal search around the predicted point within a small range to find matching feature points. The algorithm can avoid traverse the whole image to search feature points, greatly reduce the consumption of feature point matching process time.

## II. MATERIALS AND METHODS

In the study of Visual-SLAM algorithm, feature point matching is a very important process [17] [18]. The image feature point matching process firstly extracted the feature points of each image captured by the camera, and match with the adjacent frame image to find the location of each feature point in the adjacent frames and preserve these locations. Next, through each feature point position in the current image and adjacent image, plus the camera parameters, calculate the coordinate's change of each feature in three-dimensional coordinates. It is assumed that the intrinsic parameters of camera are previously estimated [19]. Further get attitude and displacement changes of the camera.

However, this process of matching is quite time consuming, in the environment of the high real-time demand, especially in embedded systems, it's hard to do real-time SLAM work. This paper put forward a kind of algorithm that introduced inertial measurement unit data into the image feature point matching process, predict feature point location, so that can reduce the time of feature points matching. Prediction algorithm will first extract the current frame image and the next frame image feature points and record the location and descriptor of the feature points. Then using the position of feature in the current frame and inertial measurement unit data, predict the position of the feature points in the next frame. In the feature point matching process, only match feature points near the predicted position. If successful match, record the position of the matching feature points matching success; if the match fails, give up match this feature point, continue to match the next feature point immediately. When all the feature points matching process completed, record all the position information of feature points matching success.

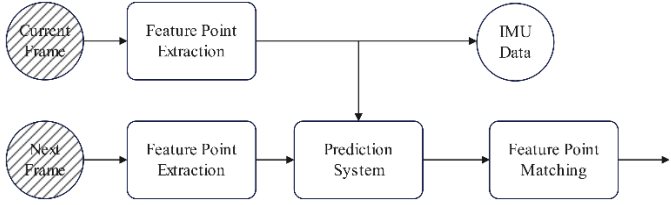

Figure 1. System Diagram.

## III. PROPOSED PREDICTION ALGORITHM

Our algorithm devotes to predicting feature point position in the next frame, thus reducing time consuming when traverse whole image to match feature point. Algorithm consists of the following steps: first, the feature points extraction algorithm is used to find feature points in two adjacent images and save feature point coordinates and descriptor. Then using inertial measurement data to calculate the camera's status changes, include displacement changes and angle changes; The state changes of the camera will be next substituted into the prediction system to calculate the positions of the feature points in the next frame; Finally, neighboring matching only do feature point matching around the prediction location. Feature point extraction algorithm is not our focus, this paper no longer gives unnecessary detail. The following focuses on introduce camera status solving process, prediction system and the neighboring matching algorithm.

### A. Camera Status Solving Process

Camera status solving process is the preparation of prediction system, it can provide camera's position changes and angle changes for prediction system. Install a gyroscope and an accelerometer on the camera, which can measure the instantaneous moment of the angular velocity and acceleration value. The direction of the angular velocity is $r$, $p$ and $y$; the direction of the acceleration is $x$, $y$ and $z$. First, reading the value of the gyroscope and accelerometer at regular intervals $\Delta t$ to get the angular velocity $\omega$ and acceleration $a$ of the camera at this moment. $\omega$ and $a$ vector shown below:

$$\omega = [\omega_r, \omega_p, \omega_y] \quad (1)$$

$$a = [a_x, a_y, a_z] \quad (2)$$

If the time interval selected $\Delta t$ is very small, this paper assumes that in $\Delta t$ angular velocity and acceleration is constant. Multiplying the angular velocity and acceleration with time can get the change value of angle $\Delta \theta$ and velocity $\Delta v$ during $\Delta t$:

$$\Delta \theta = [w_r, w_p, w_y] \cdot \Delta t \quad (3)$$

$$\Delta v = [a_x, a_y, a_z] \cdot \Delta t \quad (4)$$

Then, using the last moment angle value $\theta_L$ plus angle changed value $\Delta \theta$ can get the current angle $\theta_C$ of camera:

$$\theta_C = \theta_L + \Delta \theta \quad (5)$$

Also, using the last moment velocity value $v_L$ plus velocity changed value $\Delta v$ can get the current velocity $v_C$ of camera:

$$v_C = v_L + \Delta v \quad (6)$$

To calculate camera's displacement, using the following (7) and (8):

$$\Delta r = v_L + \frac{1}{2} \cdot [a_x, a_y, a_z] \cdot \Delta t \quad (7)$$

$$r_C = r_L + \Delta r \quad (8)$$

In the above equation, $\Delta r$ represent camera's displacement during the interval $\Delta t$, $r_L$ is the displacement of the camera relative to the initial position (t = 0) at the last moment, and $r_C$ is the displacement of the camera this moment.

One thing needs to notice is that when record data, not only to record every moment value of camera's angle $\theta_C$ and displacement $r_C$, but also record the corresponding time $t$ from starting time, the three data is very important in the prediction system.

### B. Prediction system

Prediction system is responsible for predicting the location of the feature points in the next frame. Prediction system firstly project the feature points to camera 3D coordinate based on the plane homograph. Then according to the change of camera's angle $\Delta \theta$ and displacement $\Delta r$ which get in "Camera status solving process" section, calculate rotation matrix $R$ and displacement matrix $r$. Multiplying feature points' 3D coordinates with $R$ and plus $r$ can get a new set of 3D coordinates, which is the predicted positions' 3D coordinates in the next camera state. Finally, Prediction system project predicted positions' 3D coordinates to 2D coordinate and get the predicted position in the next frame. Next, with a feature point $p_1(u_1, v_1)$ matching process as an example, the following will detail system prediction steps.

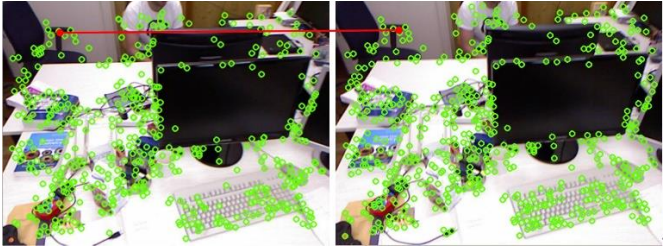

Figure 2. Two adjacent frame. The left is previous frame and the right is current frame. The green circles represent the feature points; red line represents two feature point were matched.

The RGB-D camera which RGB-D SLAM used adopt pinhole camera model. A pixel in camera coordinate defined as $[x, y, z]$, and in frame coordinate defined as $[u, v, d]$ ($d$ represents depth data), these two definitions has corresponding relationship shown as following:

$$u = \frac{x \cdot f_x}{z} + c_x$$

$$v = \frac{y \cdot f_y}{z} + c_y \qquad (9)$$

$$d = z \cdot s$$

In (9), $f_x$ and $f_y$ refers to the focal length of the camera in the X axis and Y axis respectively, $C_x$ and $C_y$ refers to the camera's aperture center, s refers to the scaling factor of the depth image. Using the (9) can project current frame's feature point $p_1(u_1, v_1)$ to camera's 3D coordinate $P_1(x_1, y_1, z_1)$.

As introduced in "camera status solving process" section, the algorithm is recording every moment value of camera's angle $\theta_c$ and displacement $r_c$, as well as the corresponding time $t$. When prediction system calculates rotation matrix $R$ and displacement matrix $r$, it requires two group elements: camera's angle $\theta_1$ and displacement $r_1$ at the time $t_1$ when camera capture current frame; and angle $\theta_2$ and displacement $r_2$ at the time $t_2$ when camera capture next frame. Put $r_1$ and $r_2$ into (8) can get displacement matrix $r$:

$$r = r_2 - r_1 \qquad (10)$$

The rotation angle of the camera from time $t_1$ to time $t_2$ can be expressed by the following (11):

$$[\psi, \theta, \varphi] = [\theta_{2z}, \theta_{2y}, \theta_{2x}] - [\theta_{1z}, \theta_{1y}, \theta_{1x}] \qquad (11)$$

$[\psi, \theta, \varphi]$ are respectively rotate angle around x axis, y axis and z axis.

$$q = \begin{bmatrix} w \\ x \\ y \\ z \end{bmatrix} = \begin{bmatrix} \cos(\varphi/2)\cos(\theta/2)\cos(\psi/2) + \sin(\varphi/2)\sin(\theta/2)\sin(\psi/2) \\ \sin(\varphi/2)\cos(\theta/2)\cos(\psi/2) - \cos(\varphi/2)\sin(\theta/2)\sin(\psi/2) \\ \cos(\varphi/2)\sin(\theta/2)\cos(\psi/2) + \sin(\varphi/2)\cos(\theta/2)\sin(\psi/2) \\ \cos(\varphi/2)\cos(\theta/2)\sin(\psi/2) - \sin(\varphi/2)\sin(\theta/2)\cos(\psi/2) \end{bmatrix} \qquad (12)$$

Using (12) to redefine $[\psi, \theta, \varphi]$ by quaternions format.

$$R = \begin{bmatrix} 1 - 2q_2^2 - 2q_3^2 & 2q_1q_2 + 2q_0q_3 & 2q_1q_3 - 2q_0q_2 \\ 2q_1q_2 - 2q_0q_3 & 1 - 2q_1^2 - 2q_3^2 & 2q_2q_3 + 2q_0q_1 \\ 2q_1q_3 + 2q_0q_2 & 2q_2q_3 - 2q_0q_1 & 1 - 2q_1^2 - 2q_2^2 \end{bmatrix} \qquad (13)$$

Rotation matrix $R$ can calculate by (13).

$$[x, y, z]^T = R \times [x_1, y_1, z_1]^T + r \qquad (14)$$

Put $P_1(x_1, y_1, z_1)$ into (14) can get a three-dimensional coordinate point $P(x, y, z)$. Using (9), projecting $P$ as a two-dimensional coordinate point $p(u, v)$, get the feature point $p_1$'s predicted position $P$ in next frame.

## C. Neighboring Matching

After frame feature point extraction process, can get two frame's feature points set $P_n = \{p_1, p_2, ..., p_3\}$ and $Q_m = \{q_1, q_2, ..., q_m\}$. Traditional matching algorithm will take out feature point $p_n$ from $P$ in order, then traversal $Q$ to find is their matched feature point $q_m$ for $p_n$. This means that matching all points must perform $n \times m$ times matching algorithm. However, using prediction system can obtain $p_1$'s predicted position $p(u, v)$ from prediction system. When traversing $q_m$, if $q_1$ is not near $p(u, v)$, shows that $q_1$ is probably not the $p_1$'s matched feature point, so there is no need to perform matching algorithm, skip $q_1$, verify whether $q_2$ near $p(u, v)$. Not until find a $q_m$ near the coordinates $p(u, v)$ did system perform matching algorithm. Neighboring matching algorithm process is shown in the following pseudo code.

| Neighboring matching algorithm |
| --- |
| For q1: qm |
|   If u-threshold<q1.u<u + threshold |
|     If v-threshold<q1.v<v + threshold |
|       Feature Match; |
|     Else break |
|   Else break |

According to $p_1$'s predicted position $p(u, v)$ can only matching feature points around coordinates $p(u, v)$ in the next frame. So, the time of matching algorithm is far less than $n \times m$ times. Experiments show that the threshold selected 10 pixels is more appropriate.

## IV. RESULTS AND DISCUSSION

To test the predictive performance of algorithm, this paper evaluates it from two aspects: time consuming and error rate. The platform used to do experiment is a Turtlebot robot with Kinect. Turtlebot robot provide IMU information and Kinect collect images. The benchmark datasets used in the experiment is initial, front1m, back1m, left30, and right30 which means robot at initial point, move forward 1m, move back 1m, turn left 30 degrees and turn right 30 degrees respectively. This paper selects SIFT, SURF, ORB and FAST as contrast algorithm to extract and match feature points, and the program of these algorithm come from OpenCV 2.4.13 function library. This paper compares the optimized algorithm with the SIFT, SURF, ORB and FAST algorithm from two aspects: Time consuming and Error rate. The output of our algorithm is shown in Figure 3.

## A. Time Consuming

This paper randomly selects two images which at an interval of 2 seconds from freiburg1_xyz, freiburg1_360 and freiburg2_360_kidnap database, as shown in Figure 3. First, using traditional SIFT, SURF, ORB and FAST algorithm to match feature points for these three groups image, and obtain the time consuming before optimized. After that, embedded predict system in the traditional algorithm program, getting optimized time consuming. This time includes the time of reading RGB image and depth image, feature point extraction and feature points matching. In order to ensure that experimental results are rigorous, only increase prediction algorithm in traditional SIFT, SURF, ORB and FAST algorithm program, which means optimized algorithm and traditional algorithm program are basically same.

From the experimental results, the average time consuming: ORB is 70.231ms, SIFT is 9307.996ms, SURF is 315.88ms, FAST is 149.631ms and our algorithm is 69.121ms. Comparing with SIFT, our algorithm's time consume is reduced more than 12 times. Because when matching image feature points, the traditional algorithm need traverse the whole image to find the matching feature point. But the optimized algorithm doesn't need to traverse the whole image for matching feature points, just around the predicted point within a small range to find matching feature points, so the optimized algorithm can save much time.

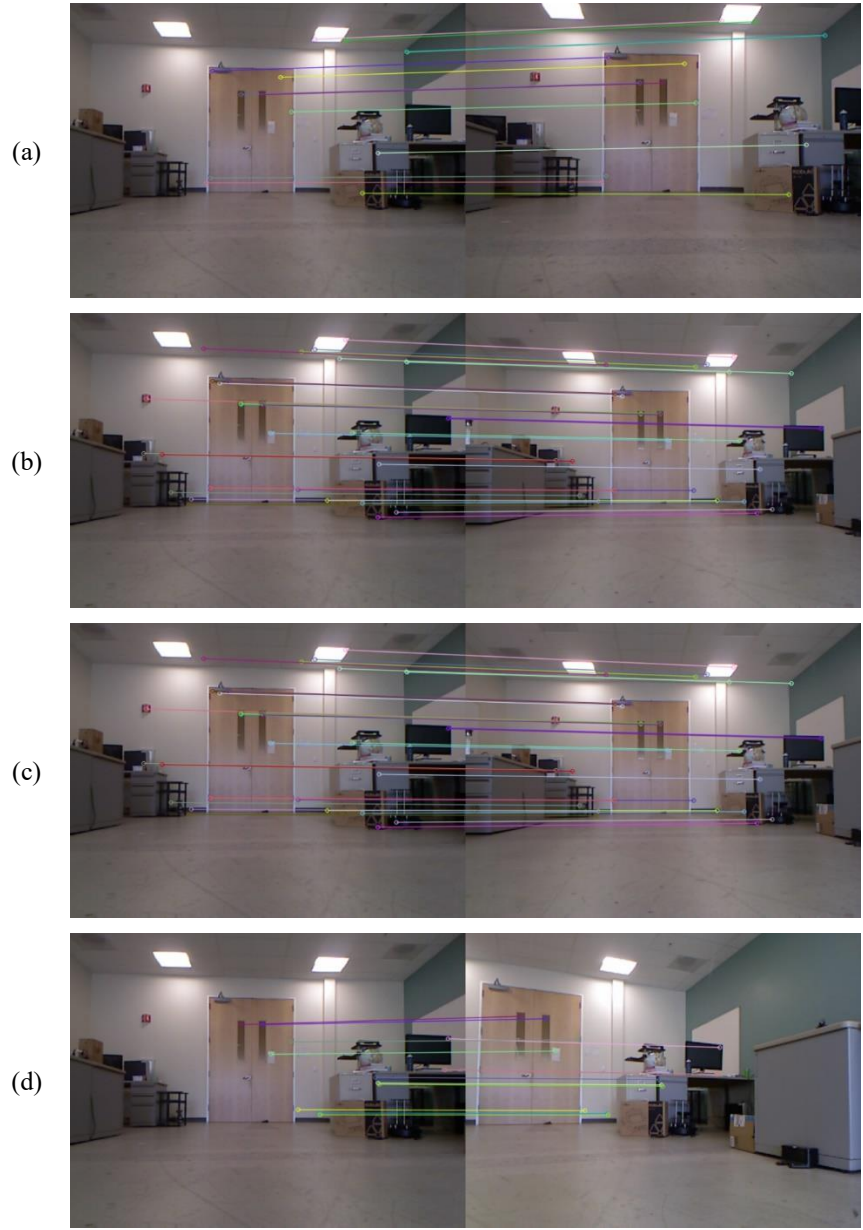

Figure 3. Tested frame. (a) initial point compares with font1m; (b) initial point compares with back1m; (c) initial point compares with left30; (d) initial point compares with right30.

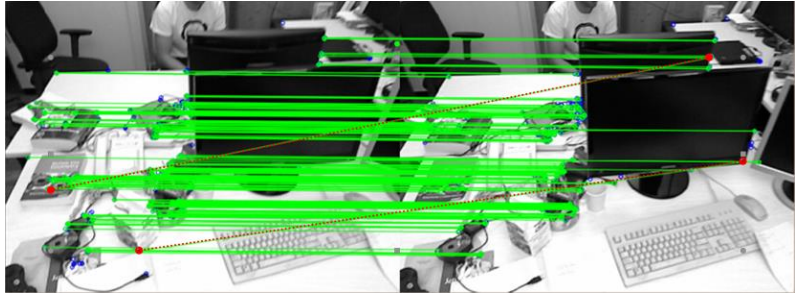

Figure 4. False matches are marked with red dashed lines

TABLE I. TIME CONSUMING FOR FEATURE MATCH

|  | SIFT | SURF | FAST | ORB | Our algorithm |
|---|---|---|---|---|---|
| Front1m | 6552.635 | 145.620 | 104.332 | 39.356 | **40.856** |
| Back1m | 8235.368 | 341.309 | 150.997 | 81.379 | **80.677** |
| Left30 | 14523.506 | 495.597 | 202.564 | 104.860 | **100.042** |
| Right30 | 7920.473 | 280.993 | 140.632 | 55.327 | **54.908** |
| Average time | 9307.996 | 315.88 | 149.631 | 70.231 | **69.121** |

*B. Error Rate*

For the four sets of images in Figure 4, this paper first statistic the total number of matched feature points and the number of false matched feature points using each algorithm. Then calculate the average total number and the average false number of features for each algorithm. Result is shown in Table 2.

TABLE II. STATISTICAL RESULT

|  | SIFT | SURF | FAST | ORB | Our algorithm |
|---|---|---|---|---|---|
| Average total number | 298 | 180 | 80 | 82 | **25** |
| Average false number | 30 | 19 | 10 | 5 | **0** |

The results show that the improved algorithm can effectively reduce the number of errors, and in some cases, the number of errors even reduced to 0. Because the advantage of prediction algorithm, the optimized feature algorithm points only match feature point around the predicted position. Therefore, it can prevent the occurrence of error matching in a certain extent, such as situation shown in Figure 4.

## V. CONCLUSION

This paper focus on the problem of reducing time consuming of feature points matching in visual SLAM system. First, discuss the problem that traditional matching algorithm takes considerable time. The reason is that traditional matching algorithm need to traverse all the second image's feature points when matching the feature points of two image. Each feature point needs to perform matching algorithm. This paper put forward a method of using camera's angle and displacement change to predict feature point's location in next frame. The prediction position can provide a reference coordinate for feature point matching process. Feature point whose coordinates are not near the predicted location may not be the right feature points, so do not need to perform the matching algorithm.

Results on open dataset show that prediction algorithm can reduce the time-consuming of feature point matching process and improve the accuracy of the feature point matching. Algorithm optimized are also inexpensive in computation cost. In RGB-D SLAM, the Iterative Closest Point (ICP) algorithm also takes a long time. And the depth value of the RGB-D camera is often invalid or of high uncertainty, which lead to the result of the ICP is not accurate. Using prediction algorithm in ICP algorithm to reduce time consuming and improve accuracy, will be future work.